# A Contrastive Learning Framework for Breast Cancer Detection

**Samia Saeed [1, 2] and Khuram Naveed [2, 3] ***

1    Department of Electrical Engineering, University of Wah (UW), WahCantt; samia.saeed@wecuw.edu.pk.
2    Department of Electrical and Computer Engineering, COMSATS University Islamabad (CUI), Islamabad, Pakistan;
3    Department of Dentistry and Oral Health, Aarhus University Denmark; knaveed@dent.au.dk.
*    Correspondence: knaveed@dent.au.dk, samiasaeed0006@gmail.com;

**Abstract:** Breast cancer, the second leading cause of cancer-related deaths globally, accounts for a quarter of all cancer cases [1]. To lower this death rate, it is crucial to detect tumors early, as early-stage detection significantly improves treatment outcomes. Advances in non-invasive imaging techniques have made early detection possible through computer-aided detection (CAD) systems which rely on traditional image analysis to identify malignancies. However, there is a growing shift towards deep learning methods due to their superior effectiveness. Despite their potential, deep learning methods often struggle with accuracy due to the limited availability of large-labeled datasets for training. To address this issue, our study introduces a Contrastive Learning (CL) framework, which excels with smaller labeled datasets. In this regard, we train Resnet-50 in semi supervised CL approach using similarity index on a large amount of unlabeled mammogram data. In this regard, we use various augmentation and transformations which help improve the performance of our approach. Finally, we tune our model on a small set of labelled data that outperforms the existing state of the art. Specifically, we observed a 96.7% accuracy in detecting breast cancer on benchmark datasets *INbreast* and *MIAS*.

## 1. Introduction

Breast cancer is the transformation of the breast's healthy cells into tumor cells, typically through a multi-stage process that advances from pre-cancerous lesions to malignant masses. This rapid proliferation of abnormal cells extends beyond their original confines and infiltrate neighboring organs ultimately leading to the demise of the patient. Unfortunately, of all documented cancer cases, Breast Cancer with a ratio of 25% [2] is second in mortality after lung cancer. Nonetheless, early detection coupled with an appropriate treatment regimen typically results in a favorable prognosis for it. In this setting, Computer-Aided Diagnosis (CAD) tools play a crucial role. These systems interpret patient scans to find soft markers related to both benign and cancerous structures, even before symptoms appear, making them indispensable in the diagnostic process.

With the emergence of advanced imaging modalities, a range of methods for detecting breast anomalies has been studied. Initially, conventional methods such as mathematical morphology [3] and fractal dimension analysis [4] were tested. These techniques, while foundational, displayed partial efficacy, especially in finding intricate features like micro-calcifications and masses across diverse cases. The limits of classical image processing paved the way for autonomous systems, augmented by the integration of Machine Learning (ML) algorithms. Weighted Bayesian classifiers [5], SVM classifiers [6], and Random Forest methods [7] were among the first ML techniques employed for breast cancer detection. Yet, their dependence on manually curated features and domain-specific regulation limited their generalizability and performance. Ultimately, the revolution came with the rise of Deep Learning (DL) practices, which reformed the field by empowering systems to learn subtle patterns directly from the data. Convolutional Neural Networks (CNNs) [8], in particular, appeared as a keystone in this revolution. These architectures, designed to mimic the human visual cortex's hierarchical processing, established substantial steps in accuracy. CNN variants [9], including architectures like ResNet50 [10], proved especially proficient at addressing issues such as the vanishing gradient problem in deep networks. By deploying deep residual learning, ResNet50, and similar models enhanced the robustness of feature extraction and upgraded the general performance metrics. Despite the patent accomplishments of DL systems, the landscape remains challenging. One of the most persistent matters is the need for large-scale, accurately labeled datasets to efficiently train these sophisticated models. This prerequisite highlights the requirement for an association between medical experts and technologists to constantly improve and adapt these technologies for sophisticated diagnostic procedures.

In short, while the evolution from conventional image processing to ML and DL has upgraded early-stage detection of breast cancer, thus dropping death rates, these methods still face constraints due to the shortage of annotated data that is vital to empower them. To this end, we aim to address the challenge of low breast cancer detection accuracy due to labeled data scarcity through a Self-supervised (SSL) Contrastive Learning framework, as it has the potential to overcome the restraints for small-scale datasets. Unlike supervised learning which depends on labeled samples, SSL techniques like Contrastive Learning (CL) extract meaningful features from data without explicit labels. CL [11] precisely functions by comparing representations of correlated views of the same data point based on a similarity measure i.e. it finds and reinforces similarities between positive pairs (similar samples) while pushing apart negative pairs (dissimilar samples) in the embedded space. This approach helps apprehend primary features and disparities within the data, thus enhancing the model's ability to generalize across different cases. In contrast to generative or discriminative SSL approaches, which either generate (pseudo-)labels or learn discriminative boundaries, CL focuses on identifying the intrinsic patterns and associations within data samples. This method is predominantly valued in medical imaging, where labeled datasets are often limited and costly.

## 2. Materials and Methods

The methodology addresses accuracy challenges in cancer detection caused by a shortage of labeled data through a Contrastive Learning (CL) framework. This includes extracting features using a ResNet50 with an MLP and using the Contrastive Loss function to discriminate data classes by adjusting the distances between similar and dissimilar points. Succeeding the pre-training phase, the model is tweaked with a small labeled dataset to enhance the performance and lastly, the model's classification decision is studied using GradCam [12].

**1**. *Transformation module (T)* at the start of the framework generates two augmented versions of an input image (x), denoted as (xi and xj), by applying random augmentations (t and t'). These augmentations are based on empirical observations rather than clinical standards, which is crucial due to the complex variability and intricate features of medical imaging, including contrast, blur, noise, artifacts, and distortion. In contrastive learning, where the aim is robust representation learning, it is important to use a variety of augmentations simultaneously rather than relying on just one method. This approach helps capture disease-specific features located in small, critical areas and improves the accuracy and reliability of diagnostic interpretations. The following augmentations are implemented to achieve this:

a. *Random Flip*: Flipping generates a mirror image of an image with either horizontal or vertical axes. Random flipping is employed as image parts are interchangeable in the case.

b. *Random Crop*: The images are randomly cropped from 50% to 80%. Cropped versions of images can help deep neural networks learn features regardless of their original location. Furthermore, the range applied determines the safety of this augmentation technique by preserving the structure of image along with the regions of interest.

c. *Resize*: Images are then resized to 32x32. Increasing image resolution for training often has a trade-off with the maximum possible batch size, yet the selection of higher image resolution has the potential for further increasing neural network performance [11].

d. *Color jitter*: In a grayscale medical image, the hue (apparent color shade) and saturation (apparent color intensity) of each pixel is equal to 0. The lightness (apparent brightness) is the only parameter of a pixel that can vary from black to white with a range of grayscale. For augmentation, we have first used brightness to obtain a true black that will maximize contrast, and then adjust contrast with 20% probability [11].

e. *Gaussian blur*: Gaussian blur is not used to examine fine details of the images, which otherwise might get harder [13].

After normalization, the pipeline independently processes each image twice, yielding two views (xi and xj), presumed to form a positive pair. No explicit negative sampling is conducted rather a mini-batch (MB) produces positive pairs, with the remaining data points treated as negative samples. The images then undergo feature extraction through the backbone model.

**2**. *The backbone model* comprises two main components: a base encoder and a projection head. Within the Contrastive Learning (CL) framework, the base encoder is a Convolutional Neural Network (CNN) that handles augmented views (xi and xj) to extract underlying features (hi and hj), which are then mapped to a representation in real latent space ($IR_d$). To overcome problems such as vanishing gradients in CNNs, ResNet-50 [10], a 50-layer variant with skip connections as in Figure A1, is employed as the base encoder. ResNet-50 [10] is recognized for its efficacy in deep learning due to its capacity to sustain low error rates even in deeper networks. It accomplishes this through skip connections that connect every three layers and direct networks across all layers, permitting activations to drift more efficiently through the network and preserving learnings in deeper layers. This strategy makes ResNet-50 [10] predominantly effective for tasks like radiology scans, where extracting intricate features from images is vital. In terms of customizations, the model uses a 3x3 kernel size and a stride of 1x1, while replacing the max-pooling layer with identity to preserve spatial information in the images. The input is processed through four layers to produce a 2048-dimensional feature map. The classification block is removed, and a Multilayer Perceptron (MLP) is added on the top to process the resulting feature vector with ReLU activation as shown in Figure 1. This produces a non-linear representation of the feature map from the base encoder, reducing its dimensionality to 128.

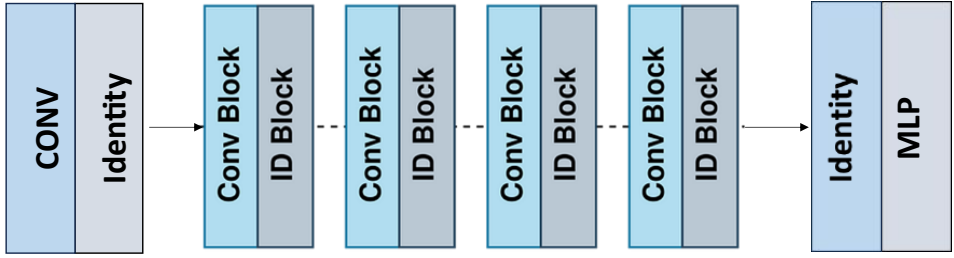

Figure 1. An illustration of the employed ResNet50

**3.** *Contrastive Loss* eventually addresses a metric-learning problem for the representation vectors from the projection head. Instead of relying on pixel-wise differences, which are ineffective due to the complex pixel interactions in images, cosine similarity is used to measure the compatibility between vectors. This is because cosine similarity, based on angular distance, aligns well with how image features are compared in vision tasks. The contrastive loss function is then computed based on cosine similarity and backpropagated as the model is trained to cluster similar images together while separating dissimilar ones as shown in Figure 2. Here temperature ($\tau$) controls the similarity function's sensitivity, thereby regulating the attraction-repulsion radius around samples. The final loss considers the similarity between all positive pairs in the set, highlighting the overall contrastive loss for the batch.

$$\text{Sim}(z_i, z_j) = z_i \cdot z_j \tag{1}$$

$$l(i,j) = \frac{\exp\left(\frac{\text{Sim}(z_i, z_j)}{\tau}\right)}{\sum_{k \neq i} \exp\left(\frac{\text{Sim}(z_i, z_j)}{\tau}\right)} \tag{2}$$

$$\frac{1}{2N}\left(\sum_{k=1}^{K} l(2k-1, 2k) + l(2k, 2k-1)\right) \tag{3}$$

**4.** *A linear classifier* is appended to the network once the base model has been pre-trained. This classifier consists of a simple linear layer that computes a weighted sum of the features extracted by the pre-trained model, followed by a softmax function (or another activation function) to produce class probabilities. The combined model, now consisting of the pre-trained layers plus the linear classifier, is then fine-tuned using a smaller labeled dataset. During this fine-tuning phase, the weights of the linear classifier are adjusted to fit the specific classification task, while the weights of the pre-trained encoder are frozen. This process helps the model adapt the learned features to the new, labeled task. After fine-tuning, the model can classify new data by passing it through the pre-trained layers to extract features, and then through the linear classifier to make final predictions as depicted in Figure 3.

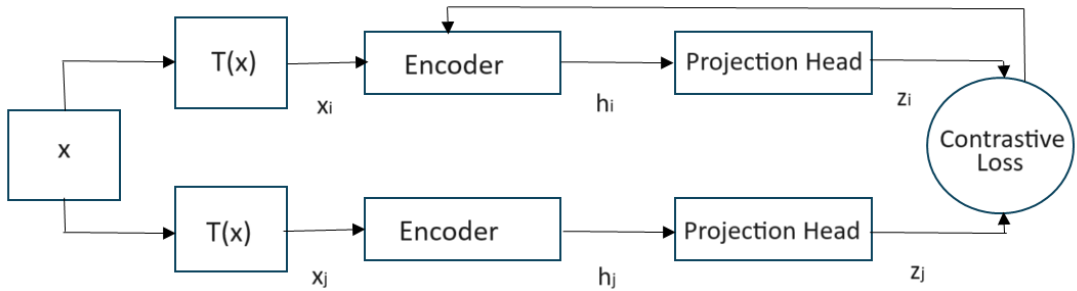

Figure 2. Block diagram of the Contrastive Learning framework

5. *Model Explanations*—additional metadata that provides insights into the model's internal workings through visual aids enhance transparency in the decision-making process. GradCam [12], a well-known visualization technique, is used to create class-specific heatmaps from the last convolutional layer. The essence of GradCam [12] is to visualize the model's classification process by leveraging the spatial information from convolutional layers. This involves computing the gradient of the predicted class with respect to the convolutional feature map, which is then averaged and pooled to assess the importance of each neuron. A weighted combination of these forward activation maps, processed by ReLU, results in the generation of the heatmap, highlighting the regions crucial for the classification decision.

## 3. Results

### 3.1 Experimental Settings

In this study, mammographic datasets are used because, among all the existing modalities mammograms are the most effective tool for early disease detection. 11277 images from "*The Digital Database for Screening Mammography (DDSM)*" [14]; a resource by the mammographic image analysis research community collected in the Breast Cancer Research Program by the U.S. Army Medical Research and Materiel Command are used without annotations in the pre-training phase of this study. Another large-scale full-field digital mammography dataset "*VinDr-Mammo*" [15] by Hanoi Medical University Hospital was also tested for pre-training. Three benchmark datasets namely "*The Mammographic Image Analysis Society (MIAS)*" [16] containing 322 images; "*INbreast*" [17] a mammographic database, with 460 4084 × 3328 images acquired at a Breast Centre, located in a Hospital de São João, Breast Centre, Porto, Portugal and 2360 images of "*King Abdul-Aziz University Mammograms (KAAUM)*" dataset [18] collected from Sheikh Mohammed Hussein Al-Amoudi Center of Excellence at King Abdul-Aziz University in Jeddah, Saudi Arabia are used for evaluation after fine-tuning. Different resolutions of the datasets are downscaled to 32x32 to make them compatible with the model.

The methodology was validated in Jupyter Notebook using a Tesla K80 GPU with CUDA Version 11.2 and a 32 GB RAM virtual machine. Hyperparameters of the framework, such as data size, batch size, and temperature of Contrastive Loss, were fine-tuned. A *LARS* optimizer optimized based on available resources with a batch size of 128 and $\tau$=0.5 was used for pre-training. Each pre-training step used a batch size of $2N$ (2*128), with one related and $2(N-1)$ disparate samples per augmented image [11]. Other hyperparameters like learning rate, weight decay, momentum, scheduling coefficient, and trust coefficient were set to conventional values. A "*Linear LR warm-up*" followed by a gradual increase to the "*main Cosine LR*" ensured effective weight optimization, maintaining regularization [11]. The model underwent 60 epochs of training, after which pre-trained model layers were frozen except the last one, which was fine-tuned using momentum-based gradient search. Classifier parameters were set to typical values, and the model was further trained for 40 epochs to refine its performance.

The network was trained in two phases: pre-training and fine-tuning. In the pre-training part, the self-supervised Contrastive Learning (CL) framework learns features from unlabeled datasets. A large batch size is utilized to increase contrastive power, managed by *Layer-wise Adaptive Rate Scaling* for stability and addressing generalization gaps. Weight decay regularization controls overfitting. The learning rate is adjusted with a "*Linear LR warm-up*" to increase it, followed by the "*main Cosine LR*" for regularization. In the fine-tuning phase, a linear classifier is employed following the pre-trained encoder, and the head's parameters are tweaked with labeled data through stochastic gradient descent. This process finds the global minimum by updating the network weights after each epoch, using randomly shuffled batches for training. Finally, the class is decided by comparing predicted class probabilities with actual targets using Cross Entropy Loss.

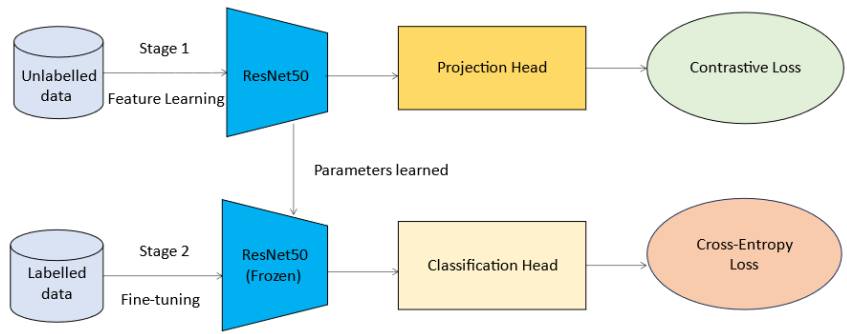

Figure 3. A depiction of the training stages

In general, the classification of tumors based on *True Positive (TP), False Positive (FP), True Negative (TN), and False Negative (FN)* are formulated to assess *accuracy, F1-measure, and AUC score*. *Accuracy* is calculated as the correct instances among total instances whereas, *F1-measure* as the ratio of correctly predicted positives with respect to total real positives. *AUC score*, is another metric that depicts the area under the *Receiver Operating Characteristic (ROC) curve*. This is a valuable metric for evaluating classification models, especially on imbalanced datasets. The model's performance was evaluated using all these measures.

$$Accuracy = \text{(TP+FP)/(TP+FP+TP+TN)} \tag{4}$$

$$F1\text{-}score = \text{2TP/(2TP+FN+FP)} \tag{5}$$

$$TPR = \text{TP/(TP+FN)} \tag{6}$$

$$AUC\text{-}score = \int\text{TPR d(FPR)} \tag{7}$$

### 3.2 Experimental Findings

The training starts by plotting the data points, which as depicted in Figure 4 are highly entangled in the latent space initially. During training, the CL framework adjusts its parameters to separate these classes in the latent space. This process involves learning representations that make the distinction between benign and malignant data more apparent. As the model trains, it refines its feature representations to reduce the overlap of the classes.

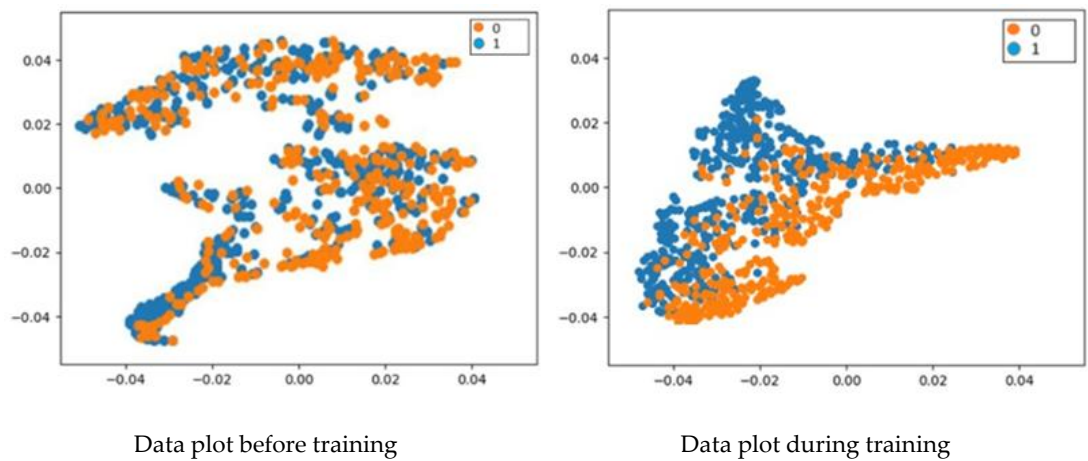

Data plot before training          Data plot during training

Figure 4. t-SNE visualizations of pre-training data

In the mid-training phase, the correct and false prediction rate of the model with pre-training on 132 images of *DDSM* [14] and fine-tuning on around 2000 images of the benchmark datasets for 45 epochs at $\tau$= 0.5 can be seen in Figure 5. As shown in (a), for *MIAS* [16], the model predicted 356 instances correctly out of 380 and for (b) *INbreast* [17] the model predicted 418 instances correctly out of 460, and for (c) *KAAUM* [18] 322 out of 380.

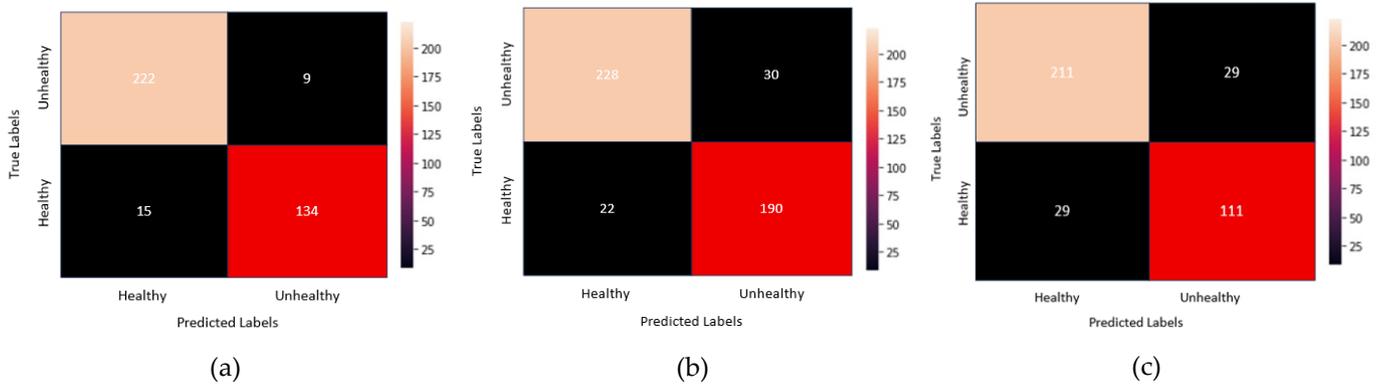

(a)                                             (b)                                             (c)

Figure 5. Confusion Matrices for benchmark datasets after 45 epochs of pre-training on (a) MIAS dataset, (b) INbreast dataset and (c) KAAUM dataset

After 60 epochs, for *MIAS* [16], *INbreast* [17] and *KAAUM* [18] datasets model has achieved reasonable *accuracy* of 95%, *F1-score* 91.02%, *AUC- score* 93.01, and *accuracy* 96.7%, *F1-score* 92.78%, *AUC-score* 94.02%, and *accuracy* 95%, *F1-score* 90.3%, *AUC-score* 92.3% respectively with the stated temperature and data-size as shown in Table1 where Figure 6 shows the (a) loss graph and (b) accuracy graph for the best-performing *INbreast* dataset [17] at 60 epochs when pre-trained on the selected settings. As stated earlier, this performance will generally get more refined by training the model for a greater number of epochs because Contrastive Learning inherently benefits from more epochs.

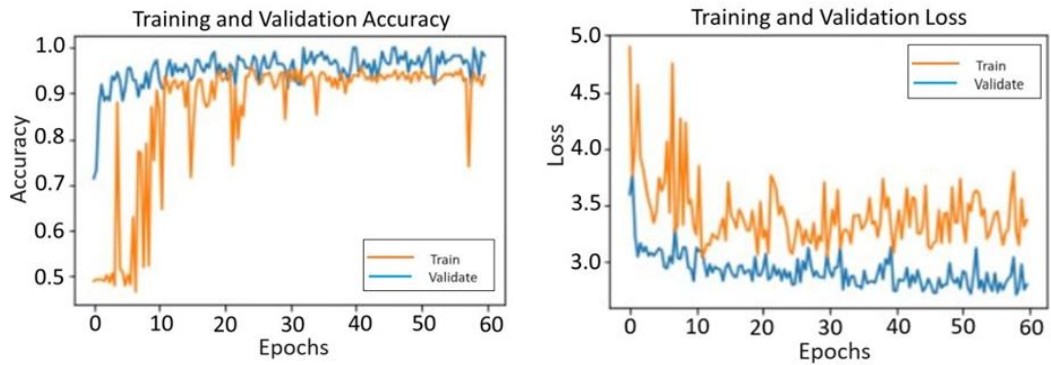

Figure 6. Proposed model's classification performance for 60 epochs, (a) Accuracy Plot and (b) Loss Plot

Finally, GradCam [12] as shown in Figure 7 is employed to visualize classification decisions. For each instance, Grad-CAM reveals which areas of the image are most influential in the model's classification thereby, for three instances, revealing the regions with significant gradients that activate their respective class activation.

Table 1. Model's performance on benchmark datasets after 60 epochs

| Dataset | Accuracy | F1-Measure | AUC-Score |
|---|---|---|---|
| *MIAS* [16] | 95 | 91.02 | 93.01 |
| *INbreast* [17] | 96.7 | 92.78 | 94.02 |
| *KAAUM* [18] | 95 | 90.3 | 92.3 |

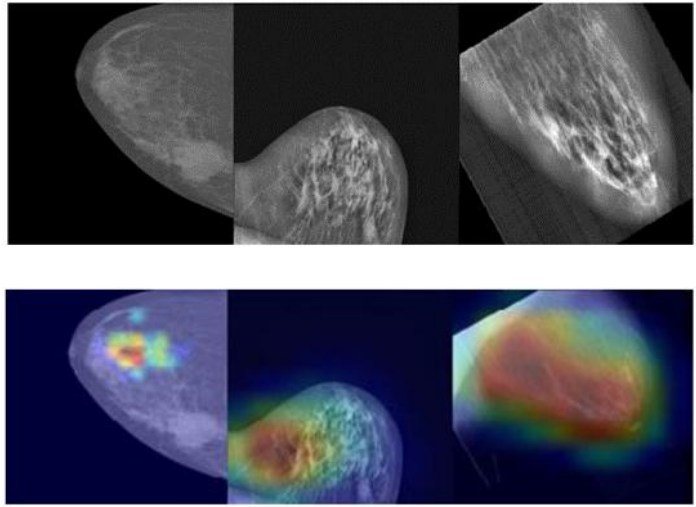

Figure 7. GradCam visualizations of the classification decision

## 4. Discussion

As the quality of representations learned is dependent on the quality of pre-training dataset therefore, we tested three different datasets to make use of the unlabeled dataset as accurately as possible. As a baseline, we first trained the model in a supervised fashion for which the model achieved the maximum accuracy of 85.% for the benchmark dataset *KAAUM* [18]. Model was then pre-trained on *VinDr* [15], *KAAUM* [18], *DDSM* [14] and it was found that it performed the best for *DDSM* [14] with an accuracy of 96.7% and 95% for benchmark *INbreast* [17] and *MIAS* [16] datasets respectively, whereas *VinDr* [15] and *KAAUM* [18] yielded an *accuracy* of 78%, 87.5% and 91%, 93% respectively with same amount of data & at the same number of epochs. Moreover, it was found that the performance of the model improves by increasing the number of unlabeled data size thereby yielding an accuracy of 98.4% on *INbreast* [17] for the pre-training data count of 44708 with half the number of epochs. Increasing the unlabeled data beyond the mentioned limit had however no effect on the training results. As Contrastive Learning benefits from a greater number of epochs and as our goal is to approach a reasonable accuracy with limited data, therefore this accuracy can be achieved by training it for a few more epochs with the earlier stated data size of 11277. Like pre-training data count, fine-tuned data also has an impact on the performance. While tuning it, we found that the performance degrades sharply if the number of images is decreased to less than 15% of the pre-training data. Then two different temperatures including $\tau = 0.1$ and $\tau = 0.5$ were tested, for which the pertaining loss can be seen in Fig.12 where (a) Contrastive Loss for $\tau = 0.5$ and (b) Contrastive Loss for $\tau = 0.1$. Though the pre-training loss was less for $\tau = 0.1$, the classification accuracy was better for $\tau = 0.5$. This relates to the fact the representations are discriminated with higher class separation for $\tau = 0.1$, but the features of $\tau = 0.5$ are more useful for the classification task, and therefore $\tau = 0.5$ was used for training purpose.

The quantitative analysis of the proposed method when done with the recently studied approaches shows a reasonable performance of the proposed work compared to other recent techniques. It can be seen in the tables where Table 2 and Table 3 compare the performance of the model with other Deep Learning techniques for benchmark *MIAS* [16] and *INbreast* datasets [17] respectively. The proposed framework, despite the minimal annotations is performing well, because CL aims to learn representations that are invariant to certain transformations (like different viewpoints or augmentations of the same image), making them more robust and generalizable. In contrast, traditional methods may struggle with generalization when faced with varied or noisy data. Moreover, traditional ML and DL methods suffer from performance degradation when the distribution of the test data differs significantly from the training data. CL, by learning a more invariant representation, is mitigating this issue to some extent.

Table 2. Comparison of the model's performance with recent studies on the MIAS dataset

| Approach | F1-Score | AUC-score | Accuracy |
|---|---|---|---|
| CNN with SVM [19] | | 93 | 93.35 |
| Homology based ML [20] | | 95 | 94.30 |
| Screen Filmed Mammograms [21] | | 99 | 84.45 |
| MIL [22] | - | 91 | 86 |
| Inception ResNet [23] | - | - | 95.83 |
| **Proposed Model** | 92.78 | 94.02 | 96.7 |

Table 3. Comparison of the model's performance with recent studies on the INbreast dataset

| Approach | F1-Score | AUC-score | Accuracy |
|---|---|---|---|
| Whale optimized CNN [24] | | | 95.46 |
| k-Means [25] | 89 | - | 90 |
| AlexNet [26] | | | 91 |
| Adaboost [27] | | 55.9 | 87.93 |
| ResNet-50 [28] | | - | 93.0 |
| **Proposed Model** | 91.02 | 93.01 | 95 |

## 5. Conclusions

This study employs small-scale labeled datasets for breast cancer detection, demonstrating that when insufficient labeled data is available, existing unannotated resources can be leveraged to achieve the objective with performance comparable to supervised models. The Self-supervised Contrastive Learning framework developed is therefore a significant step towards integrating Deep Learning into medical diagnostics. To address any discrepancies in results, it's important to note that differences may arise due to the computational resources needed to generate all possible pairs. In each dataset, there are numerous negative pairs that already meet the contrastive requirements, which can slow down training convergence. Additionally, using various data augmentation techniques alongside the proposed model can further improve outcomes, eventually reducing the gap compared to fully supervised models.

## Appendix A

ResNet-50 [10] is a deep neural network with residual learning to address issues like vanishing gradients. The core of ResNet-50 [10] is these residual blocks, each with 1x1 convolution to reduce dimensionality, 3x3 convolution to extract features, and 1x1 convolution to restore dimensionality. These blocks further have skip connections that bypass the convolutions to help address the vanishing gradient problem by permitting gradients to flow directly through the network.

The network begins with a 7x7 convolution layer featuring 64 filters and a stride of 2, succeeded by batch normalization and ReLU activation to introduce non-linearity. This is proceeded by a 3x3 max pooling layer that condenses the spatial dimensions of the feature maps, leveraging the representation compact and computationally efficient. Followed by it is the network core which unfolds systematically through four stages, each methodically arranged to augment feature extraction. This core begins with stage 1, where a single residual block lays the ground for initial feature mining followed by three additional blocks, each utilizing 64 filters, refining the feature map. As we move to stage 2 and stage 3, the sophistication ramps up with four blocks employing 128 filters and six blocks with 256 filters respectively, enabling the network to mine more subtle patterns within the data. Lastly, stage 4 concludes with three blocks equipping 512 filters finalizing the features for the final classification. This progressive assembly allows the network to advance from elementary feature extraction to complex pattern recognition, with each stage relying upon the previous to realize a refined comprehension of the input data. After these stages, a global average pooling layer reduces the feature maps into a single vector, which is then processed by a fully connected layer. This final layer employs a softmax function to generate probabilities for each class yielding the final classification.